\documentclass[sigconf]{acmart}

\def\BibTeX{{\rm B\kern-.05em{\sc i\kern-.025em b}\kern-.08emT\kern-.1667em\lower.7ex\hbox{E}\kern-.125emX}}

\usepackage{balance}
\usepackage{multirow}
\usepackage{booktabs}
\usepackage{verbatim}
\usepackage{graphicx}
\usepackage{subfigure}
    
\setcopyright{rightsretained}
\begin{document}

\fancyhead{}


\copyrightyear{2019}
\acmYear{2019}
\acmConference[MM '19]{Proceedings of the 27th ACM International Conference on Multimedia}{October 21--25, 2019}{Nice, France}
\acmBooktitle{Proceedings of the 27th ACM International Conference on Multimedia (MM '19), October 21--25, 2019, Nice, France}
\acmPrice{15.00}
\acmDOI{10.1145/3343031.3350929}
\acmISBN{978-1-4503-6889-6/19/10}

\title{Editing Text in the Wild}

%

\author{Liang Wu}
\authornote{Equal contribution. This work was mainly done when Liang Wu was an intern at Baidu Inc.}
\affiliation{%
  \institution{Huazhong University of Science and Technology}
}
\email{liangwu1995@gmail.com}


\author{Chengquan Zhang}
\authornotemark[1]
\affiliation{
  \institution{Department of Computer Vision Technology (VIS), Baidu Inc.}
}
\email{zhangchengquan@baidu.com}

\author{Jiaming Liu}
\authornotemark[1]
\affiliation{
  \institution{Department of Computer Vision Technology (VIS), Baidu Inc.}
}
\email{liujiaming03@baidu.com}


\author{Junyu Han}
\affiliation{
  \institution{Department of Computer Vision Technology (VIS), Baidu Inc.}
}
\email{hanjunyu@baidu.com}

\author{Jingtuo Liu}
\affiliation{
  \institution{Department of Computer Vision Technology (VIS), Baidu Inc.}
}
\email{liujingtuo@baidu.com}

\author{Errui Ding}
\affiliation{
  \institution{Department of Computer Vision Technology (VIS), Baidu Inc.}
}
\email{dingerrui@baidu.com}

\author{Xiang Bai}
\authornote{Corresponding author.}
\affiliation{%
  \institution{Huazhong University of Science and Technology}
}
\email{xbai@hust.edu.cn}

%
\renewcommand{\shortauthors}{Wu and Zhang, $et$ $al.$}

%
\begin{abstract}
In this paper, we are interested in editing text in natural images, which aims to replace or modify a word in the source image with another one while maintaining its realistic look. This task is challenging, as the styles of both background and text need to be preserved so that the edited image is visually indistinguishable from the source image. Specifically, we propose an end-to-end trainable style retention network (SRNet) that consists of three modules: text conversion module, background inpainting module and fusion module. The text conversion module changes the text content of the source image into the target text while keeping the original text style. The background inpainting module erases the original text, and fills the text region with appropriate texture. The fusion module combines the information from the two former modules, and generates the edited text images. To our knowledge, this work is the first attempt to edit text in natural images at the word level. Both visual effects and quantitative results on synthetic and real-world dataset (ICDAR 2013) fully confirm the importance and necessity of modular decomposition. We also conduct extensive experiments to validate the usefulness of our method in various real-world applications such as text image synthesis, augmented reality (AR) translation, information hiding, etc.

\end{abstract}

%
%

%
\keywords{Text Editing; Text Synthesis; Text Erasure; GAN}

%

%

\maketitle
\begin{figure}
\centering
\includegraphics[scale=0.5]{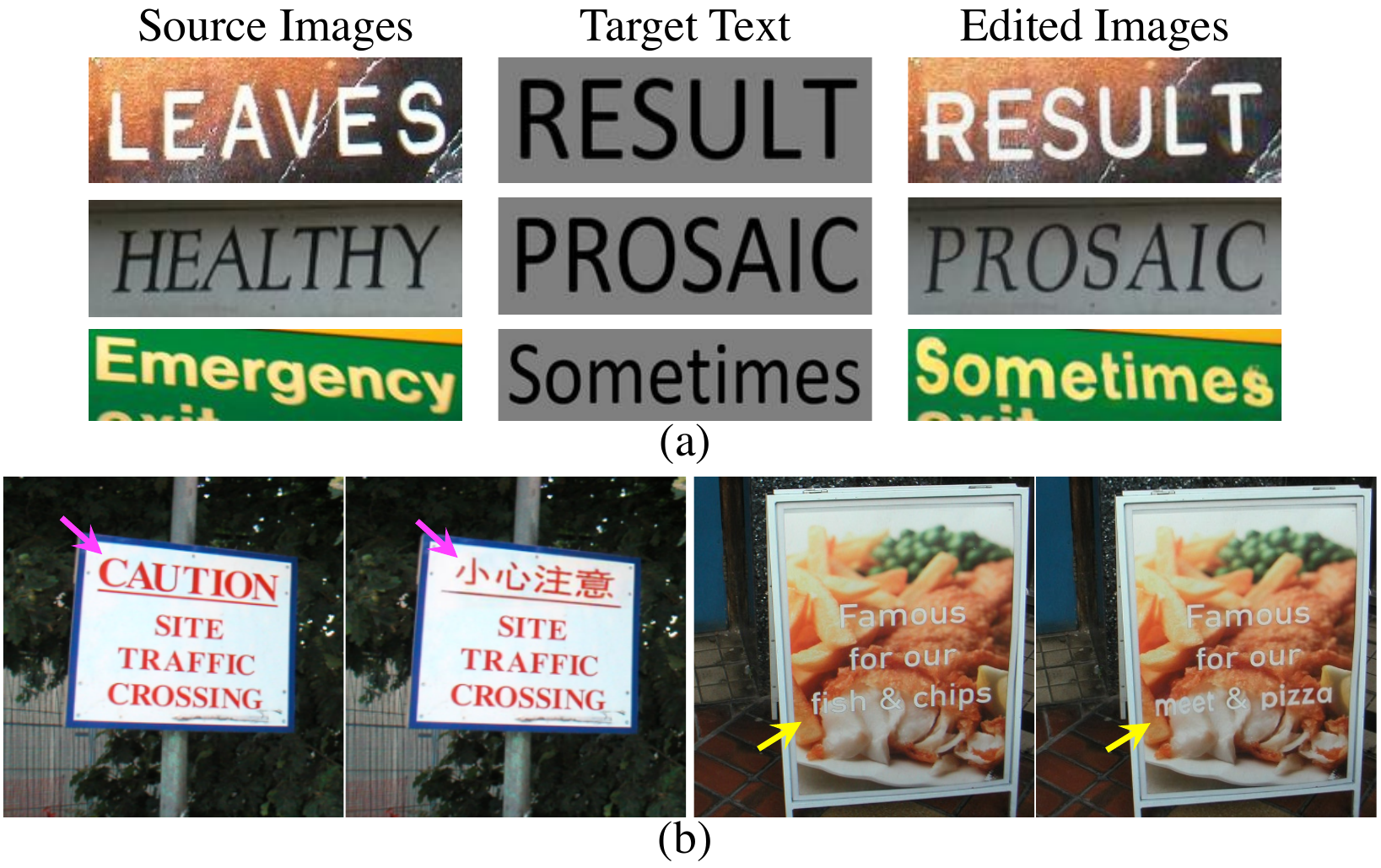}
\caption{(a) The process of scene text editing. (b) Two challenges of text editing: rich text style and complex background.}
\label{fig:intro}
\vspace{-0.4cm}
\end{figure}
\section{Introduction}
\label{sec:intro}
Text in images/videos, or known as scene text, contains rich semantic information that is very useful in many multi-media applications. In the past decade, scene text reading and its application have witnessed significant progresses~\cite{long2018scene,shi2017end,zhang2016multi,fang2018attention,zhang2019look}. In this paper, we focus on a new task related to scene text: scene text editing. Given a text image, our goal is to replace the text instance in it without damaging its realistic look. As illustrated in Fig.~\ref{fig:intro} $(a)$, the proposed scene text editor produces realistic text images by editing each word in the source image, retaining the styles of both the text and background. Editing scene text has drawn increasing attention from both academia and industry, driven by practical applications such as text image synthesis~\cite{yang2018context}, advertising photo editing, text image correction, augmented reality translation~\cite{fragoso2011translatar}.

As shown in Fig.~\ref{fig:intro} $(b)$, there are two major challenges for scene text editing: text style transfer and background texture retention. Specially, the text style consists of diverse factors 
such as 
language, font, color, orientation, stroke size and spatial perspective, which makes it hard to precisely capture the complete text style in source image and transfer them to the target text. Meanwhile, it is also difficult to maintain the consistency of the edited background, especially when text appears on some complex scenes, such as menu and street store sign. Moreover, if the target text is shorter than the original text, the exceeding region of characters should be erased and filled with appropriate texture.

Considering these challenges, we propose a style retention network (SRNet) for scene text editing which learns from pairs of images.
The core idea of SRNet is to decompose the complex task into several simpler, modular and joint-trainable sub networks: text conversion module, background inpainting module and fusion module, as illustrated in Fig.~\ref{fig:arch}. Firstly, the text conversion module (TCM) transfers the text style of the source image to the target text, including font, color, position, and scale. In order to keep the semantics of the target text, we introduce a skeleton-guided learning mechanism to the TCM, whose effectiveness has been verified in Exp.~\ref{exp:ablation}. At the same time, the background inpainting module (BIM) erases the original text stroke pixels and fills them with appropriate texture in a bottom-up feature fusion manner, following the general architecture of a "U-Net"~\cite{ronneberger2015u}. Finally, the fusion module automatically learns how to fuse foreground information and background texture information effectively, so as to synthesize edited text image.

Generative Adversarial Networks (GAN) models~\cite{goodfellow2014generative,isola2017image,zhu2017unpaired} have achieved great progress in some tasks, such as image-to-image translation, style transfer, these methods typically apply the encoder-decoder architecture that embeds the input into a subspace then decodes it to generate desired images. Instead of choosing such a single branch structure, the proposed SRNet decomposes the network into modular sub networks, while decomposes the complex task into several easy-to-learn tasks. This strategy of network decomposing has been proven useful in recent works~\cite{andreas2016learning,balakrishnan2018synthesizing}. Besides, the experiment results of SRNet are better than pix2pix~\cite{isola2017image}, a successful method used in image-to-image translation, which further confirms the effectiveness and robustness of SRNet. Compared with the work of character replacement~\cite{roy2019stefann}, our methods works in a more efficient word-level editing way. In addition to the ability to edit scene text image in the same language (such as the English words on ICDAR 2013), SRNet also shows very encouraging results in cross-language text editing and information hiding tasks, as exhibited in Fig.~\ref{fig:translation},~\ref{fig:eras}.  

The major contribution of this paper is the style retention network (SRNet) proposed to edit scene text image. SRNet possesses obvious advantages over existing methods in several folds: 
\begin{itemize}
\vspace{-0.1cm}
\item[$\bullet$] To our knowledge, this work is the first to address the problem of word or text-line level scene text editing by an end-to-end trainable network; 
\item[$\bullet$] We decompose SRNet into several simple, modular and learnable modules, including a text conversion module, a background inpainting module and the final fusion module, which enables SRNet to generate more realistic results than most image-to-image translation GAN models;
\item[$\bullet$] Under the guidance of stroke skeleton, the proposed network can keep the semantic information as much as possible;
\item[$\bullet$] The proposed method exhibits superior performance on several scene text editing tasks like intra-language text image editing, AR translation (cross-language), information hiding (e.g. word-level text erasure), etc.
\end{itemize}

\section{Related Work}

\subsection{GAN}
Recently, GANs~\cite{goodfellow2014generative} have attracted increasing attention and made great progress in many fields ,including generating images from noise~\cite{mirza2014conditional}, image-to-image translation~\cite{isola2017image}, style transfer~\cite{zhu2017unpaired}, pose transfer~\cite{zhu2019progressive}, etc. 
The framework of GANs consists of two modules: generator and discriminator, where the former aims to generate data close to the realistic distribution while the latter strives to learn how to distinguish between real and fake data. DCGAN~\cite{radford2015unsupervised} firstly used convolutional neural networks (CNN) as the structures of generator and discriminator, improved training stability of GAN. Conditional-GAN~\cite{mirza2014conditional} generated the required images under the constraints of given conditions, and achieved significant results in pixel-level alignment image generation task. Pix2pix~\cite{isola2017image} implemented the mapping task from image to image, which was able to learn the mapping relationship between input domain and output domain. Cycle-GAN~\cite{zhu2017unpaired} accomplished the cross-domain conversion task under the unpaired style images while achieving excellent performance. However, existing GANs are difficult applied in text editing task directly, because the text content changes while the shape of text needs change greatly, and the complex background texture information also need to be preserved well when editing a scene text image.

\subsection{Text Style Transfer}
Maintaining the scene text style consistency before and after editing is extremely challenging. There are some efforts attempting to migrate or copy text style information from a given image or stylized text sample. Some methods focus on character-level style transfer, for example, Lyu $et$ $al.$~\cite{lyu2017auto} proposed an auto-encoder guided GAN to synthesize calligraphy images with specified style from standard Chinese font images. Sun $et$ $al.$~\cite{sun2017learning} used a VAE structure to implement a stylized Chinese character generator. Zhang $et$ $al.$~\cite{zhang2018separating} tried to learn the style transfer ability between Chinese characters at the stroke level. Other methods focus on text effects transfer, which can learn visual effects from any given scene image and bring huge commercial value in some specific applications like generating special-effects typography library. Yang $et$ $al.$~\cite{yang2017awesome,yang2018context}proposed a patch-based texture synthesis algorithm that can map the sub-effect patterns to the corresponding positions of the text skeleton to generate image blocks. It is worth noting that this method is based on the analysis of statistical information, which may be sensitive to glyph difference and thus induce a heavy computational burden. Recently, TET-GAN~\cite{yang2019tet} used the GAN to design a lightweight framework that can simultaneously support stylization and destylization on a variety of text effects. Meanwhile, MC-GAN~\cite{azadi2018multi} used two sub-networks to solve English alphabet glyph transfer and effect transfer respectively, which accomplished the few-shot font style transfer task.
\begin{figure*}[t]
    \centering
    \includegraphics[scale=0.5]{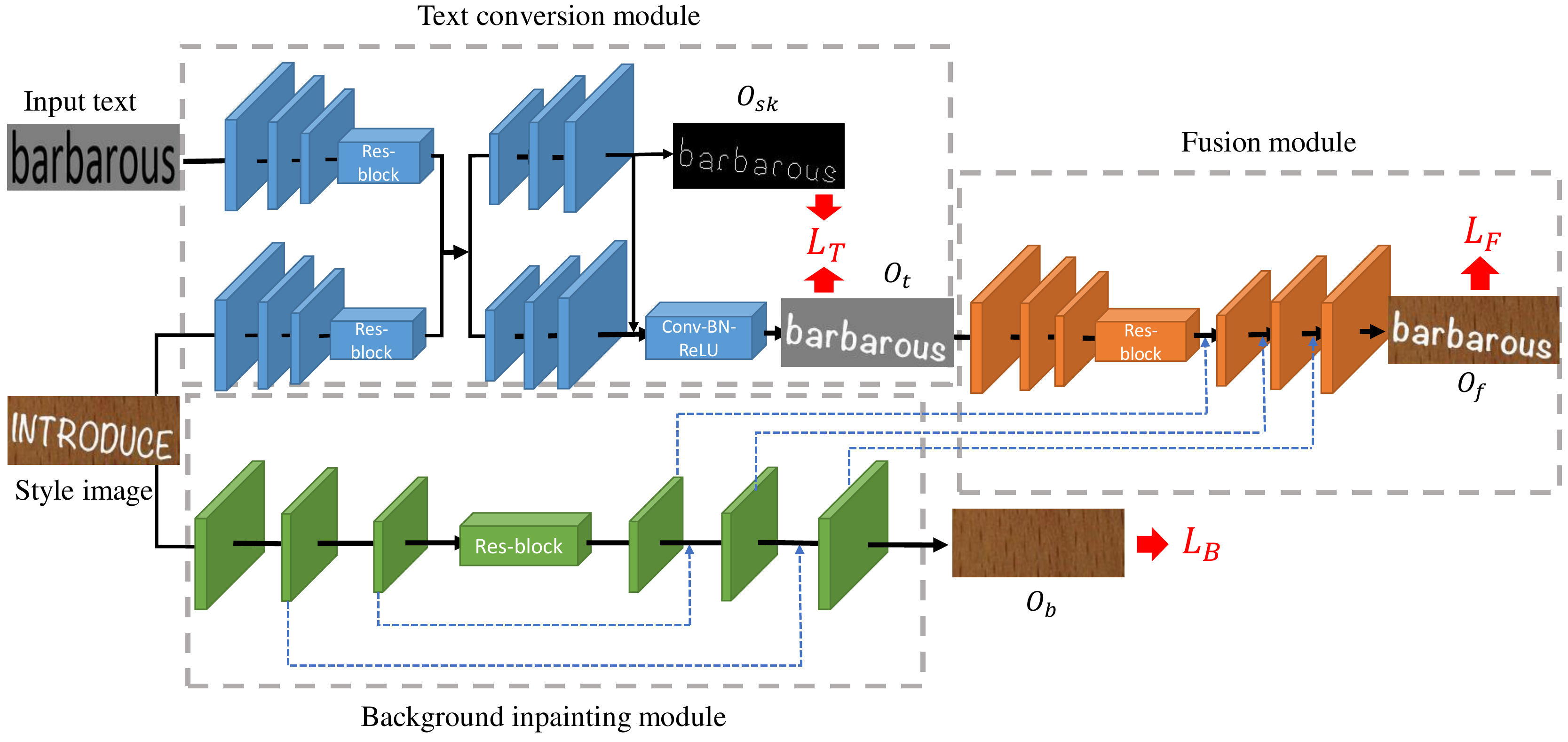}

\caption{The overall structure of SRNet. The network consists of a skeleton-guided text conversion module, a background inpainting module and a fusion module.}
\label{fig:arch}
\end{figure*}

Different from these existing methods, the proposed framework in this paper is trying to solve the migration problem of arbitrary text styles and special effects at a word or text-line level, rather than at the character level. In practice, word-level annotations are much easier to obtain than character-level annotations, and editing word is more efficient than editing characters. Besides, word-level editors favor word-level layout consistency. When dealing with words of different lengths, our word-level editor can adjust the placement of foreground characters adaptively, while character-level methods ignore.

\subsection{Text Erasure and Editing}
Background texture needs to be consistent with that before editing for scene text editing. There are some related works of text erasure, trying to erase the scene text stroke pixels while completing image inpainting on corresponding positions. Nakamura $et$ $al.$~\cite{nakamura2017scene} proposed an image-patch based framework for text erasure, but large computational cost is induced due to the sliding window based processing mechanism. EnsNet~\cite{zhang2019EnsNet} firstly introduced the generative adversarial network to text erasing, which can erase the scene text on the whole image in an end-to-end manner. With the help of refined loss, the visualization results looks better than those of pix2pix~\cite{isola2017image}. Our background inpainting module is also inspired by generative adversarial networks. In the process of text editing, we only pay attention to background erasure at word-level, therefore, the background inpainting module in SRNet can be designed more light and still have good erasure performance which is illustrated in Fig.~\ref{fig:eras}.

We noticed that a recent paper~\cite{roy2019stefann} try to study the issue of scene text editing, but it can only transfer the color and font of a single character in one process while ignoring the consistency of background texture. Our method integrates the advantages of the approaches of text style transfer and text erasing. We propose a style retention network which can not only transfer text style by an efficient manner (word or text-line level processing mechanism) but also retain or inpaint the complete background regions to make the result of scene text editing more realistic.

\section{METHODOLOGY}

We present a style retention network (SRNet) for scene text editing. During training, the SRNet takes as input a pair of images $(I_s, I_t)$ where $I_s$ is the source style image and $I_t$ is the target text image. The outputs $((T_{sk}, T_t), T_b, T_f)$ where $T_{sk}$ is the target text skeleton, $T_t$ is the foreground image which has the same text style as $I_s$. $T_b$ is the background of $I_s$ and $T_f$ is the final target text image.
In order to effectively tackle the two major challenges mentioned in Sec.~\ref{sec:intro}, we decompose the SRNet into three simpler and learnable sub networks: 1) text conversion module, 2) background inpainting module and 3) fusion module, as illustrated in Fig.~\ref{fig:arch}. Specifically, the text style from source image $I_s$ is transferred to the target text with the help of a skeleton-guided learning mechanism aiming to retain text semantics(Sec.~\ref{sec:Text Conversion Module}). Meanwhile the background information is filled by learning an erasure or inpainting task (Sec.~\ref{sec:Background Inpainting Module}). Lastly, the transferred target image and completed background are fused by the text fusion network, generating the edited image (Sec.~\ref{sec:Fusion Module}).

\subsection{Text Conversion Module}
\label{sec:Text Conversion Module}
We render the target text into a standard image with a fixed font and background pixel value setting to 127, and the rendered image is denoted as target text image $I_t$.
The text conversion module (blue part in Fig.~\ref{fig:arch}) takes the source image $I_s$ and the target text image $I_t$ as inputs, and aims to extract the foreground style from the source image $I_s$ and transfers it to the target text image $I_t$. In particular, the foreground style contains text style, including font, color, geometric deformation, and so on. Thus, the text conversion module outputs an image $O_t$ which has the semantics of the target text and the text style of the source image. An encoder-decoder FCN is adopted in this work. For encoding, the source image $I_s$ is encoded by $3$ down-sampling convolutional layers and $4$ residual blocks~\cite{he2016deep}, the input text image $I_t$ is also encoded by the same architecture, then two features are concatenated along their depth axis. For decoding, there are $3$ up-sampling transposed convolutional layers and $1$ Convolution-BatchNorm-LeakyReLU blocks to generate the output $O_t$. Moreover, we introduce a skeleton-guided learning mechanism to generate more robust text. We use $G_T$ to denote the text conversion module and the output can be represented as:
\begin{equation}
O_t = G_T(I_t,I_s). 
\end{equation}

\textbf{Skeleton-guided Learning Mechanism.} Different from other natural objects, humans distinguish different texts mostly according to the skeleton or glyph of text. It is necessary to maintain the text skeleton in $I_t$ after transferring the text style from source style image $I_s$.
To achieve this, we introduce a skeleton-guided learning mechanism. 
Specifically, we add a skeleton response block which is composed of $3$ up-sampling layers and $1$ convolutional layer followed by a sigmoid activation function to predict a single channel skeleton map, and then concatenate the skeleton heatmap and decoder output along depth axis. We use the dice loss~\cite{milletari2016v} instead of the cross-entropy loss to measure the reconstruction quality of the skeleton response map since it is found to yield more accurate results. Mathematically, the skeleton loss is defined as: 
\begin{equation}
\label{equ:dice}
\mathcal{L}_{sk} = 1 - \frac{2 \sum_i^N (T_{sk})_i (O_{sk})_i}{\sum_i^N (T_{sk})_i + \sum_i^N (O_{sk})_i},
\end{equation}
where $N$ is the number of pixell; $T_{sk}$ is the skeleton ground truth map; $O_{sk}$ is output map of the skeleton module. 

We further adopt the $L1$ loss to supervise the output of text conversion module. Combing with the skeleton loss, the text conversion loss is:
\begin{equation}
 \mathcal{L}_T = {\Vert T_t - O_t \Vert}_1 + \alpha \mathcal{L}_{sk},
\end{equation}
where $T_t$ is the ground truth of text conversion module, and $\alpha$ is regularization parameter, which is set to $1.0$ in this paper.

\subsection{Background Inpainting Module}
\label{sec:Background Inpainting Module}
In this module, our main goal is to obtain the background via a word-level erasure task. As depicted in the green part in Fig.~\ref{fig:arch}, this module takes only the source image $I_s$ as its input, and outputs a background image $O_b$, in which all text stroke pixels are erased and filled with proper texture. The input image is encoded by $3$ down-sampling convolutional layers with stride $2$ and follows with $4$ residual blocks, then the decoder generates the output image with original size via 3 up-sampling convolutional layers. We use the leaky ReLU activation function after each layer while tanh function for the output layer. We denote the background generator as $G_B$.
In order to make the visual effects more realistic, we need to restore the texture of background as much as possible.
U-Net~\cite{ronneberger2015u}, which proposes to add skip connections between mirrored layers, proven remarkably effective and robust at solving object segmentation and image-to-image translation tasks. 
Here, we adopt this mechanism in the up-sampling process, where previous encoding feature maps with the same size are concatenated to reserve richer texture. This helps to restore the lost background information during the down-sampling process.

Different from other full text image erasure methods~\cite{zhang2019EnsNet,nakamura2017scene}, our method aims at word-level image inpainting task. Text appearing in word-level image tends to be relatively standard in scale, so our network structure has possesses simple and neat design. Inspired by the work of Zhang $et$ $al.$~\cite{zhang2019EnsNet},
the adversarial learning is added to learn more realistic appearance.
The detailed architecture of the background image discriminator $D_B$ is described in Sec.~\ref{sec:Discriminators}. The whole loss function of background inpainting module is formulated as:
\begin{equation}
\begin{split}
 \mathcal{L}_B = \mathbb{E}_{(T_b,I_s)} [\log D_B( & T_b,I_s)] + \mathbb{E}_{I_s} \log [1-D_B(O_b,I_s)] + \\
 & \beta {\Vert T_b - O_b \Vert}_1,
\end{split}
\end{equation}
where $T_b$ is the ground truth of background. The formula is combined by adversarial loss and $L1$ loss, and $\beta$ is set to $10$ in our experiments.

\subsection{Fusion Module}
\label{sec:Fusion Module}
The fusion module is designed to fuse the target text image and background texture information harmoniously, so as to synthesize edited scene text image. 
As the orange part illustrates in Fig.~\ref{fig:arch}, the fusion model also follows the encoder-decoder FCN framework. 
We feed the foreground image, generated by text conversion module, to the encoder, which consists of three down-sampling convolutional layers and residual blocks. 
Next, a decoder with three up-sampling transposed convolutional layers and Convolution-Batch-Norm-LeakyReLU blocks to generates the final edited image. 
It is noteworthy that we connect the decoding feature maps of the background inpainting module to the corresponding feature maps with the same resolution in the up-sampling phase of the fusion decoder. 
In this way, the fusion network outputs the images whose background details are substantially restored; text object and background are fused well while achieving synthesis realism in the appearance. We use $G_F$ and $O_f$ to denote the fusion generator and its outputs respectively. Besides, the adversarial loss is added here, and the detailed structure of the corresponding discriminator $D_F$ will be introduced in Sec.~\ref{sec:Discriminators}. In summary, we can formulate the optimization objectives of the fusion module as the following:

\begin{equation}
\begin{split}
 \mathcal{L}_F' = \mathbb{E}_{(T_f,I_t)} [\log D_F( & T_f,I_t)]  + \mathbb{E}_{I_t} \log [1-D_F(O_f,I_t)] + \\
 & \theta_1 {\Vert T_f - O_f \Vert}_1,
\end{split}
\end{equation}
where $T_f$ is the ground truth of edited scene images. We choose $\theta_1=10$ to keep balance between adversarial loss and $L1$ loss.

\begin{figure*}[t]
    \centering
    \includegraphics[height=9cm,width=17.5cm]{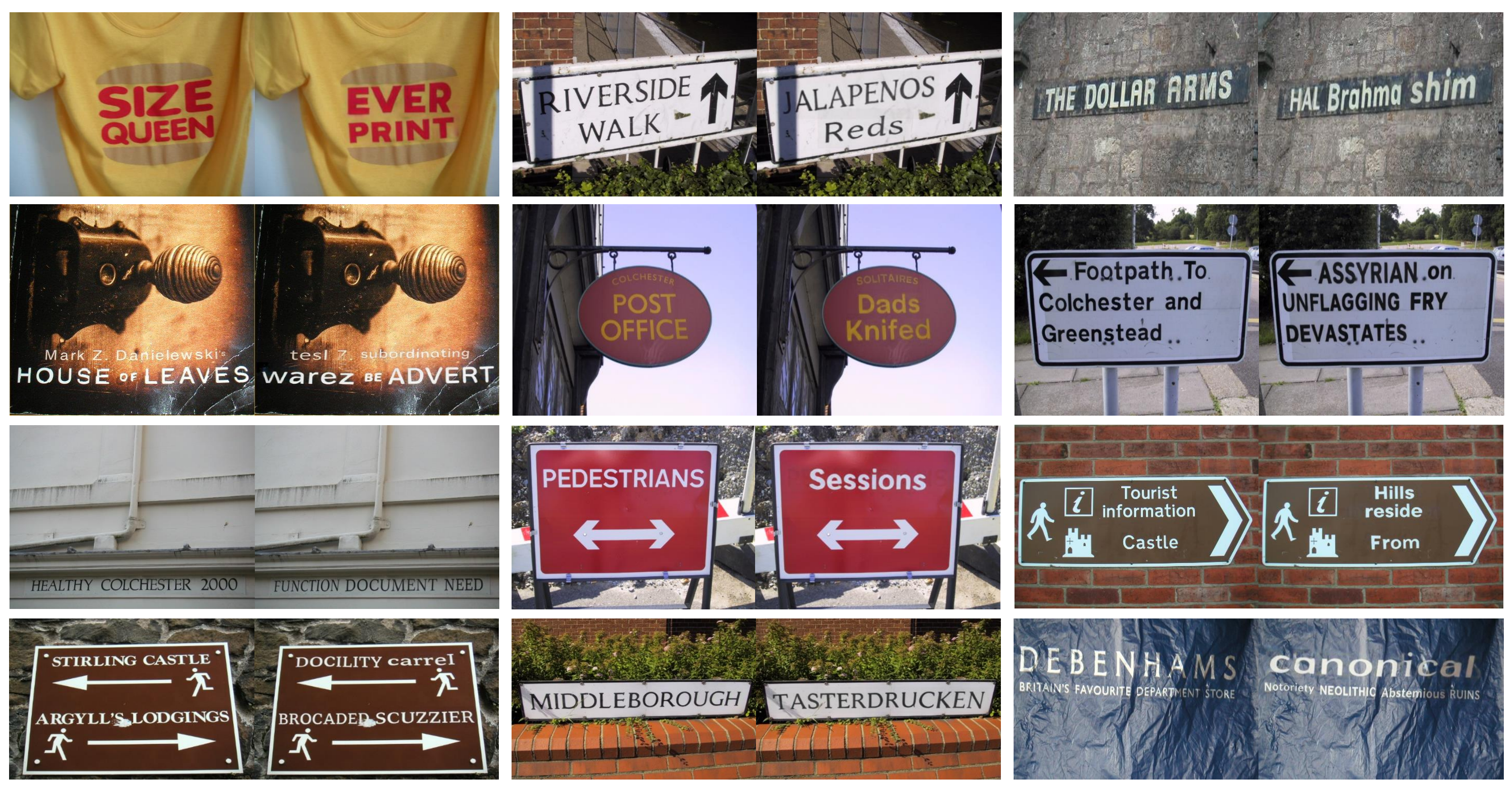}

\caption{Some results on ICDAR2013 dataset. Images
from left to right: input images and edited results. It should be noted that on the third row we replaced the words whose lengths is different from the original text; the last row shows some cases with long text.}
\label{fig:result_show}
\vspace{-0.5cm}
\end{figure*}

\textbf{VGG-Loss.} In order to reduce distortions and make more realistic images, we introduce the VGG-loss to the fusion module that includes perceptual loss~\cite{johnson2016perceptual} and style loss~\cite{gatys2016image}. As the name suggests, the perceptual loss $L_{per}$ penalizes results that are not perceptually similar to labels by defining a distance measure between activation maps of a pre-trained network (we adopt the VGG-19 model~\cite{simonyan2014very} pretrained on ImageNet~\cite{russakovsky2015imagenet}). Meanwhile, the style loss $L_{style}$ computes the differences in style. The VGG-loss $L_{vgg}$ can be represented by:
\begin{equation}
\mathcal{L}_{vgg} = \theta_2 \mathcal{L}_{per} + \theta_3 \mathcal{L}_{style},
\end{equation}
\begin{equation}
\mathcal{L}_{per} = \mathbb{E}  [\sum_i \frac{1}{M_i} {\Vert \phi_i (T_f) - \phi_i(O_f) \Vert}_1 ],
\end{equation}
\begin{equation}
\mathcal{L}_{style} = \mathbb{E}_j  [{\Vert G_j^\phi (T_f) - G_j^\phi (O_f) \Vert}_1 ],
\end{equation}
where $\phi_i$ is the activation map from relu1\_1, relu2\_1, relu3\_1, relu4\_1 and relu5\_1 layer of VGG-19 model; $M_i$ is the element size of the feature map obtained by the $i-th$ layer; $G$ is Gram matrix $ G(\textit{F}) = \textit{F}  \textit{F}^T \in \mathbb{R}^{n \times n}$; the weights $\theta_2$ and $\theta_3$ set to $1$ and $500$ respectively. The whole training objectives of the fusion model is:
\begin{equation}
 \mathcal{L}_F = \mathcal{L}_F' + \mathcal{L}_{vgg}.
\end{equation}

\subsection{Discriminators}
\label{sec:Discriminators}
Two discriminators sharing the same structure as PatchGAN\cite{isola2017image} are applied in our network. They are composed of five convolution layers to reduce the scale to 1/16 of the original size. The discriminator $D_B$ in background inpainting module concatenate $I_s$ with $O_b$ or $T_b$ as input to judge whether the erased result $O_b$ and the target background $T_b$ is similar, while the discriminator $D_F$ in fusion module concatenate $I_t$ and $O_f$ or $T_f$ to measure the consistence between the final output $O_f$ and the ground truth image $T_f$. 

\subsection{Training and Inference}
In the training stage, the whole network is trained in an end-to-end manner, and the overall loss of the model is:
\begin{equation}
\mathcal{L}_G = \arg \min_G \max_{D_B,D_F} (\mathcal{L}_T + \mathcal{L}_B + \mathcal{L}_F),
\end{equation}
Following the training procedures of GAN, we alternately train the generator and discriminators. 
We synthesize the image pairs with similar style except text as our training data. Besides, the foreground, text skeleton and background images can be obtained with the help of text stroke segmentation masks.
The generator takes $I_t$, $I_s$ as input with the supervision of $T_{sk}$, $T_t$, $T_b$, $T_f$ and outputs the text replaced image $O_t$. For the adversarial training, ($I_s$,$O_b$) and ($I_s$,$T_b$) are fed into $D_B$ to chase for background consistency; ($I_t$,$O_f$) and ($I_t$,$T_f$) are fed into $D_F$ to ensure accurate results.

In the inference phase, given the standard text image and the style image, the generator can output the erased result of style image and edited image. For the whole image, we crop out the target patches according to the bounding box annotations and feed them to our network, then we paste the results to original locations to get the visualization of whole image.

\begin{figure}[t]
    \centering

    \includegraphics[width=8.5cm]{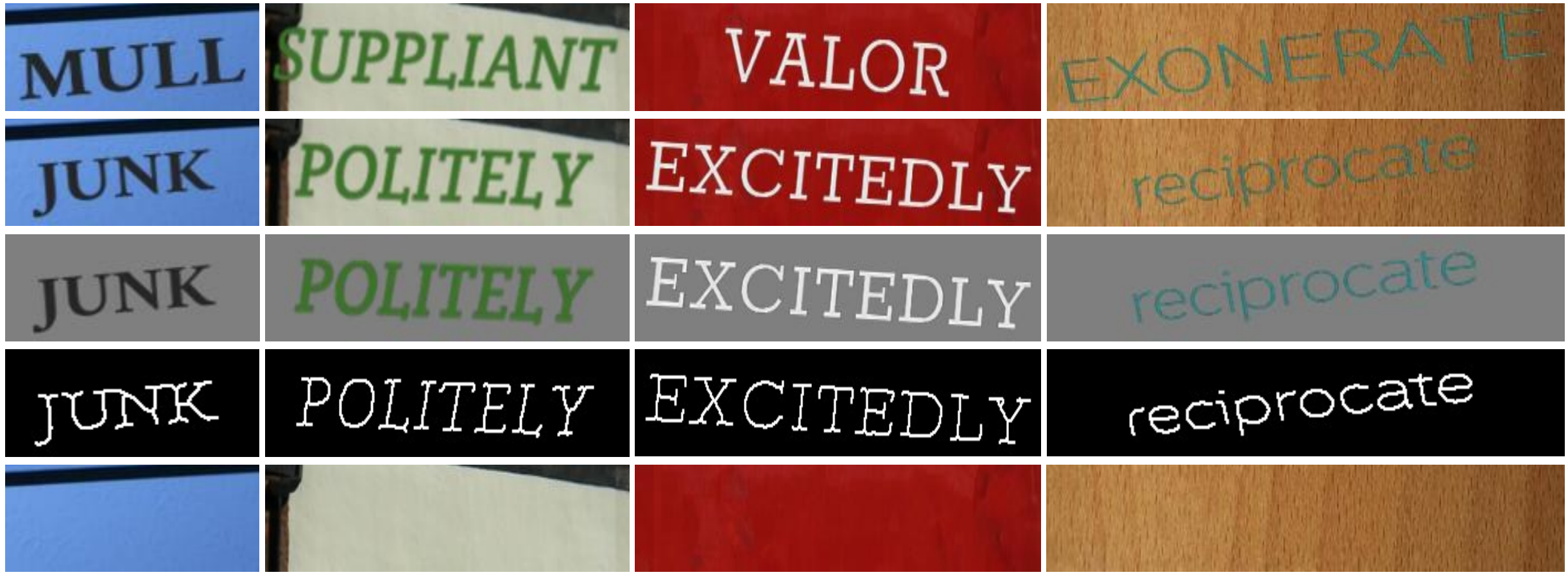}

\caption{Examples of synthetic data. From top to bottom: style image, target image, foreground text, text skeleton, background.}
\label{fig:synth}
\vspace{-0.4cm}
\end{figure}
\section{Experiments}
In this section, we present some results in Fig.~\ref{fig:result_show} to verify that our model has a strong ability of scene text editing, and we compare our method with other neural network based methods to prove the effectiveness of our approach. An ablation study is also conducted to evaluate our method.
\subsection{Datasets}
\label{sec:data}
The datasets used for the experiments in this paper are introduced as following:

\textbf{Synthetic Data}  We improve the text synthesis technology~\cite{gupta2016synthetic} to synthesize data in a pair of style but with different text, the main idea is to select fonts, color, parameters of deformation randomly to generate styled text, then render it on the background image, and, at the same time, we can get the corresponding background, foreground text and text skeleton after image skeletonization~\cite{zhang1984fast} as ground truth (Fig.~\ref{fig:synth}). In our experiments, we resize the text image height to 64 and keep the same aspect ratio. The training set consists of a total of 50000 images and the test set contains 500 images.

\textbf{Real-world Dataset} The ICDAR 2013~\cite{karatzas2013icdar} is a natural scene text data set organized by the 2013 International Conference on Document Analysis and Recognition for competition. This dataset focuses on the detection and recognition of horizontal English text in natural scenes, containing 229 training pictures and 233 test pictures. The text in each image has a detailed label and all text is annotated by horizontal rectangles. Every image has one or more text boxes. We crop the text regions according to the bounding box and input the cropped images to our network, then paste the results back to their original location. Noted that we only train our model on synthetic data, and all real-world data is used for testing only.

\subsection{Implementation Details}
We implemented our network architecture based on pix2pix~\cite{isola2017image}. Adam~\cite{kingma2014adam} optimizer is adopted to train the model with $\beta_1 = 0.5$, $\beta_2 = 0.999$ until the output tends to be stable in training phase. Learning rate is initially set to $2 \times 10^{-4}$ and gradually decayed to $2 \times 10^{-6}$ after 30 epochs. We chose $\alpha = \theta_2 = 1, \beta = \theta_1 = 10, \theta_3 = 500$ to make the loss gradient norms of each part close in back propagation. We apply spectral normalization~\cite{miyato2018spectral} to both generator and discriminator and use batch normalization~\cite{Ioffe2015Batch} in generator only. The batch size is set to 8 and the input images is resized to $w \times 64$ with the aspect ration unchanged. In training, we get the batch data randomly and the image width is resized to the average width, when testing we can input images with variable width to get desired results. The model takes about 8 hours to train with a single NVIDIA TITAN Xp graphics card.

\subsection{Evaluation Metrics} 
We adopt the commonly used metrics in image generation to evaluate our method, which includes the following: 1) MSE, also known as $\ell_2$ error; 2) PSNR, which computes the the ratio of peak signal to noise; 3) SSIM~\cite{wang2004image}, which computes the mean structural similarity index between two images. A lower $\ell_2$ error or higher SSIM and PSNR mean the results are similar to ground truth. We only calculate the above mentioned metrics on the synthetic test data, because the real dataset does not have paired data. On the real data, we calculate the recognition accuracy to evaluate the quality of the generated result. Since the input of our network is cropped image, we only compute those metrics on the cropped regions. Additionally, visual assessment is also used in real dataset to qualitatively compare the performance of various methods.

The adopted text recognition model is an attention-based text recognizer~\cite{shi2018aster} whose backbone is replaced with a VGG-like model. It is trained on Jaderberg-8M synth data~\cite{Jaderberg14c} and ICDAR 2013 training data, and them are augmented by random rotation and random resize in $x$-$axis$. 
Each text editing model renders 1000 word images based on ICDAR 2013 testing data as their respective test sets. Recognition accuracy is defined as Equ. \ref{equ:seq-acc}, where $y$ refers to the ground truth of $n-th$ sample, and $y'$ refers to its corresponding predicted result; $N$ refers to the number of samples in the whole test set. 

\begin{equation}
\label{equ:seq-acc}
seq\_acc = \frac{\sum_{n \in N_{test}}(\mathbb I(y==y'))}{N_{test}}.
\end{equation}

\begin{figure}[t]
    \centering
    \includegraphics[width=8.5cm]{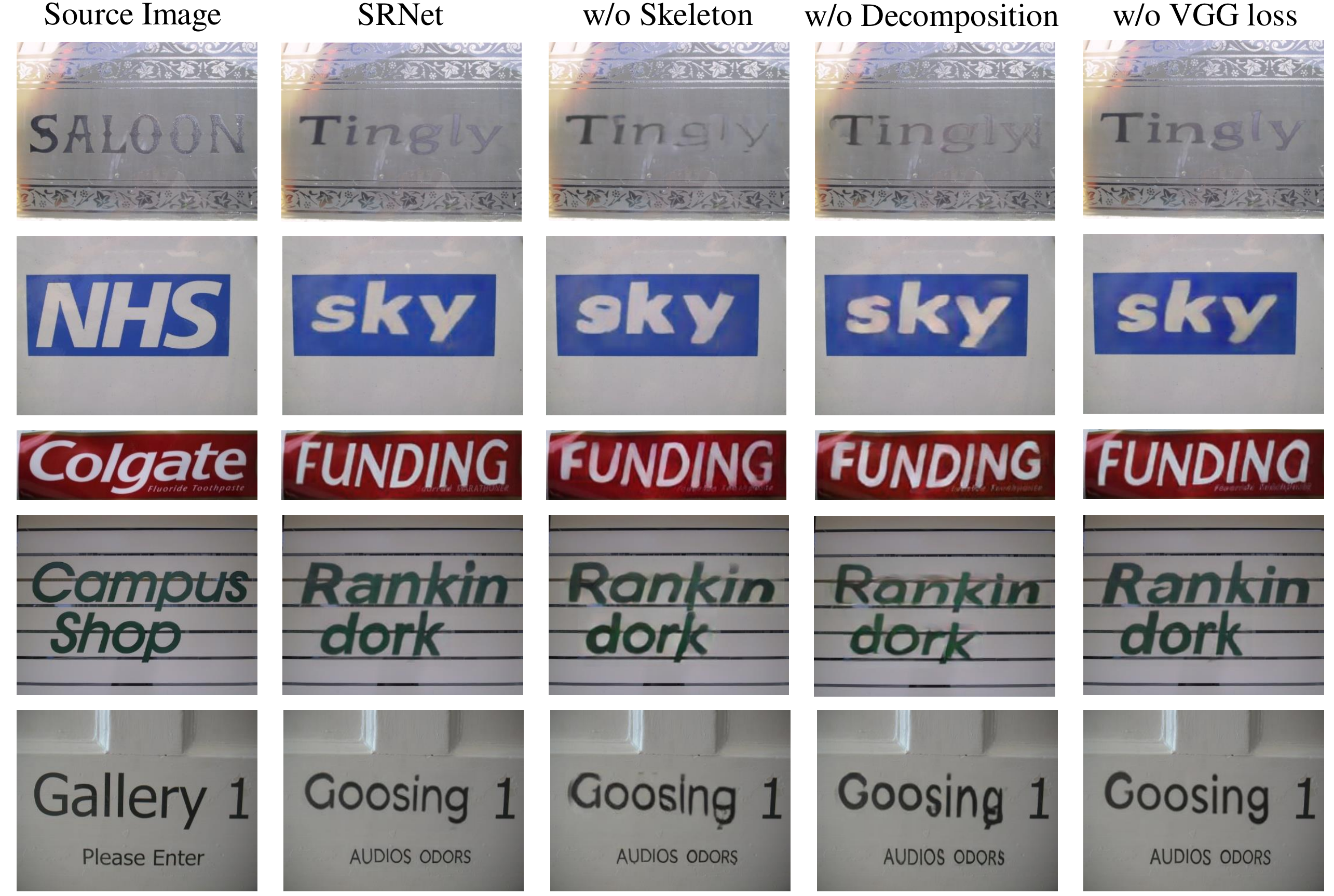}
\caption{Sample results of ablation study.}
\label{fig:ablation}
\vspace{-0.4cm}
\end{figure}

\subsection{Ablation Study}
\label{exp:ablation}
In this section, we study the effects of various components of the proposed network with qualitative and quantitative results. Fig.~\ref{fig:ablation} shows the results of different settings such as: removing the skeleton guided module, without decomposition strategy, and removing the vgg loss $L_{vgg}$(perceptual loss and style loss).

\textbf{Skeleton-guided Module.} After the removal of skeleton module, due to the lack of supervision information of the text skeleton during training, the text structure after transfer is prone to yield local bending even breakage, which is easy to affect the quality of the generated images. In contrast, the full-module method maintains the transfer text structure well and learns the deformation of the original text correctly. From Tab.~\ref{tab:metics}, we can see that the results are worse than full model on all metrics, especially a significant decline appeared in SSIM. This shows skeleton-guided module has a positive effect on the overall structure.

\begin{figure}[t]
    \centering

    \includegraphics[width=8.5cm]{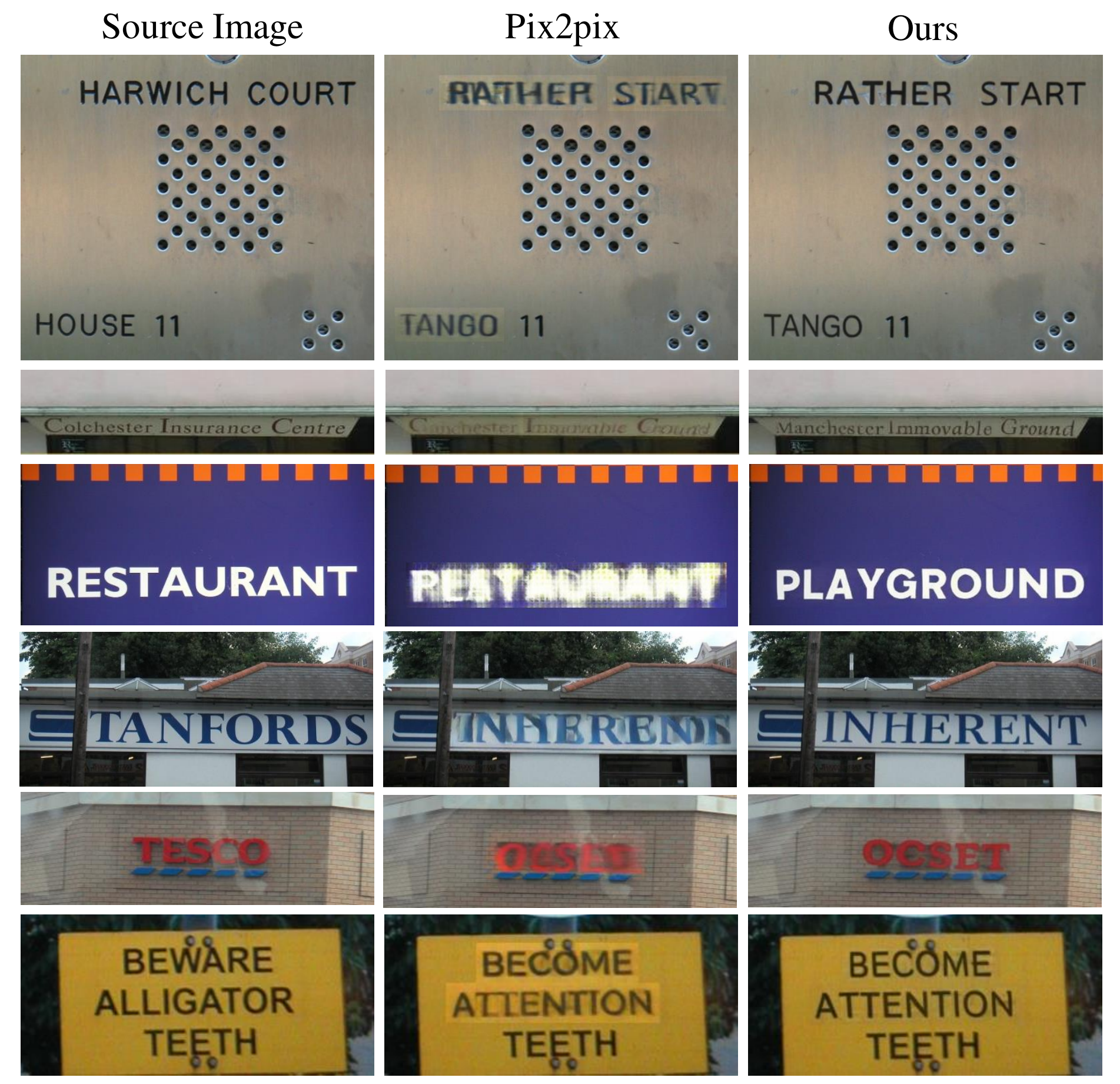}

\caption{A comparison of our model with pix2pix.}
\label{fig:comparison}
\end{figure}
\textbf{Benefits from Decomposition.} A major contribution of our work is to decompose the foreground text and background to different modules. We also conduct experiments on models that did not decompose the foreground text from background. In short, we removed the background inpainting branch, so the foreground text feature and background feature are processed by the foreground module simultaneously. From the Fig.~\ref{fig:ablation}, we can find the results are not satisfactory. The original text still remains in the synthetic image, and the text and the background are very vague. From Tab.~\ref{tab:metics}, we can find the metrics of no-decomposition are generally the worst,which verifies that the mechanism of decomposition is helpful to learn clear strokes and reduce learning complexity.

\begin{table}[]
\caption{Quantitative evaluation results.}
\setlength{\tabcolsep}{1.3mm}{
\begin{tabular}{|c|c|c|c|c|}
\hline
method                        & $\ell_2$ error    & PSNR    & SSIM   & seq\_acc    \\ \hline
pix2pix~\cite{isola2017image}                 & 0.092           & 16.54       & 0.63      & 0.717          \\ \hline
without skeleton        & 0.025           & 20.08       & 0.64      & 0.798          \\ 
without decomposition      & 0.064           & 18.56       & 0.66      & 0.786          \\ 
without vgg loss        & 0.022           & 20.39       & 0.74      & 0.778          \\ 
SRNet                    & \textbf{0.014}           & \textbf{21.12}       & \textbf{0.79}      & \textbf{0.827}          \\ \hline

\end{tabular}}
\label{tab:metics}
\vspace{-0.2cm}
\end{table}

\textbf{Discussion of VGG Loss.} As can be seen from these examples in Fig.~\ref{fig:ablation}, the results look unrealistic in appearance without the VGG loss. In this setting, we can find some details like characters in same word has different scales, the structure of text is not maintained well, etc. The results on all metrics are worse than full model, which also illustrates the importance of this component.

\begin{table}[]
\caption{Comparison SRNet with previous methods on ICDAR2013, lower value means better effect. Note that our method erased text according to the word-level annotations.}
\setlength{\tabcolsep}{3.7mm}{
\begin{tabular}{|c|c|c|}
\hline
Detection  & Erasure Methods                     & F-measure(\%) \\ \hline
           & Original image              & 75.37     \\ \cline{2-3} 
           & Pix2Pix~\cite{isola2017image}          & 17.78     \\ \cline{2-3} 
EAST~\cite{zhou2017east} & Scene text eraser~\cite{nakamura2017scene} & 16.03     \\ \cline{2-3} 
           & Ensnet~\cite{zhang2019EnsNet}            & 10.51     \\ \cline{2-3} 
           & SRNet               & \textbf{4.64}      \\ \hline
\end{tabular}}
\label{tab:eras}
\vspace{-0.2cm}
\end{table}
\subsection{Comparison with Previous Work}
Note that there was no work focusing on word-level text editing task before, so we choose pix2pix~\cite{isola2017image} network, which can complete the image translation task to compare with our method. In order to make pix2pix network implement multiple style translation, we concatenate the style image and the target text in depth as input of the network. Both methods maintain the same configurations during training. As can be seen from the Fig.~\ref{fig:comparison}, our method completes the foreground text transfer and retention of the background texture correctly; the structure of the edited text is regular; the font is consistent as before and the texture of background is more reasonable, while the results are similar to the real picture in the overall sense.
Quantitative comparison with pix2pix can be found in Tab.~\ref{tab:metics}. It indicates that our method is superior to the pix2pix method in all of the metrics.

\subsection{Cross-Language Editing}
\begin{figure}[t]
    \centering

    \includegraphics[width=8.5cm]{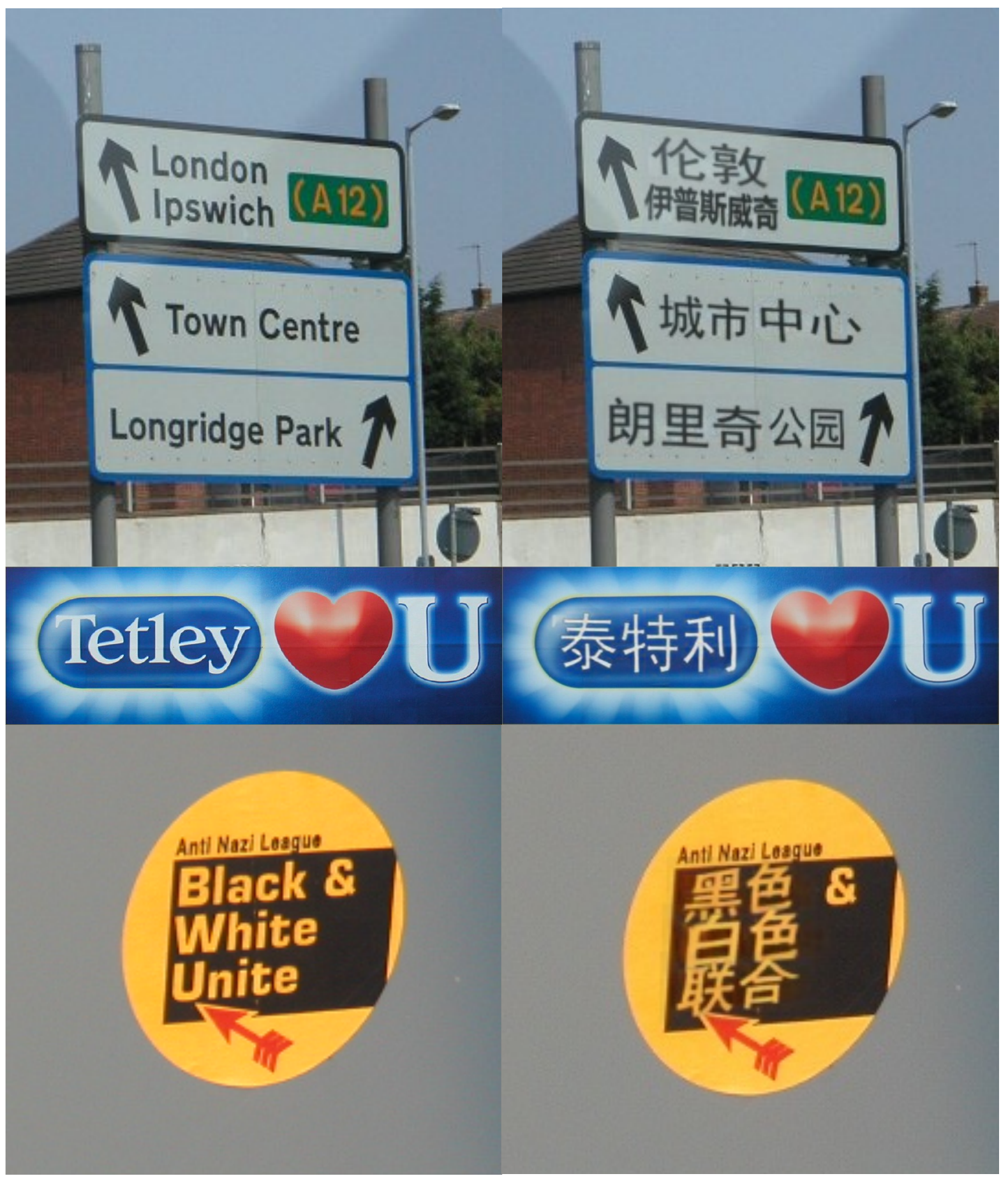}

\caption{The translation examples. Left: input images, right: translation results.}
\label{fig:translation}
\vspace{-0.1cm}
\end{figure}

In this section, we conduct an experiment on cross-language text editing task to check the generalization ability of our model. The application can be used in visual translation and AR translation to improve visual experience. Considering that the relation of Latin fonts and non-Latin fonts are not mapped well, for convenience, we only complete translation tasks from English to Chinese. 
In the training phase, we adopt the same text image synthesis method mentioned in Sec.~\ref{sec:data} to generate large amounts of training data.
It is worth noting that we map all English fonts to several common Chinese fonts manally by analyzing the stroke similarity from the size, thickness, inclination, etc.
We evaluate it on the ICDAR2013 test set and use the translation results as input text to check the generalization of our model. The results are shown in Fig.~\ref{fig:translation}, from which we can see that even if the output is Chinese characters, the color, geometric deformation and background texture can be kept very well, and the structure of characters is the same as the input text. These realistic results show the superior synthesis performance of our proposed method.

\subsection{Text Information Hiding}
\begin{figure}[t]
    \centering

    \includegraphics[width=8.5cm]{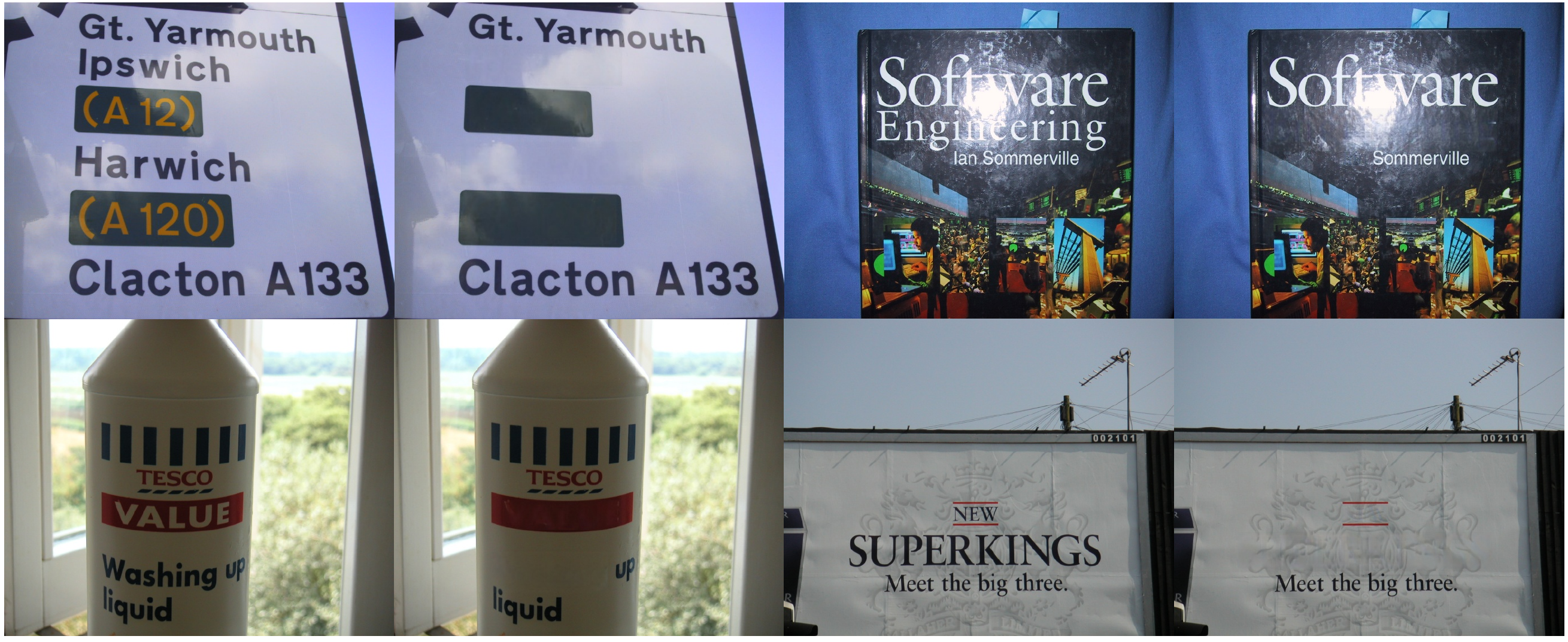}

\caption{The erasure examples. Left: input images, right: erasure results. We erase the text randomly in every image.}
\label{fig:eras}
\vspace{-0.1cm}
\end{figure}

The subtask that extracts the background information can also output the erased image. Different from the two text erasing methods~\cite{zhang2019EnsNet,nakamura2017scene}, in many cases, the entire image is not required to remove all text, it is more practical to erase part of the text in an image. We are aiming at the word-level text erasure which can select text area freely in the picture needed to be erased. As the erasure examples shown in Fig.~\ref{fig:eras}, we can see that the locations of original text are filled with appropriate textures. Tab.~\ref{tab:eras} shows the detection results on erased images. Due to the particularity of our method, we erased the cropped images and pasted them back to compare with other methods.

\begin{figure}[t]
    \centering

    \includegraphics[width=8.5cm]{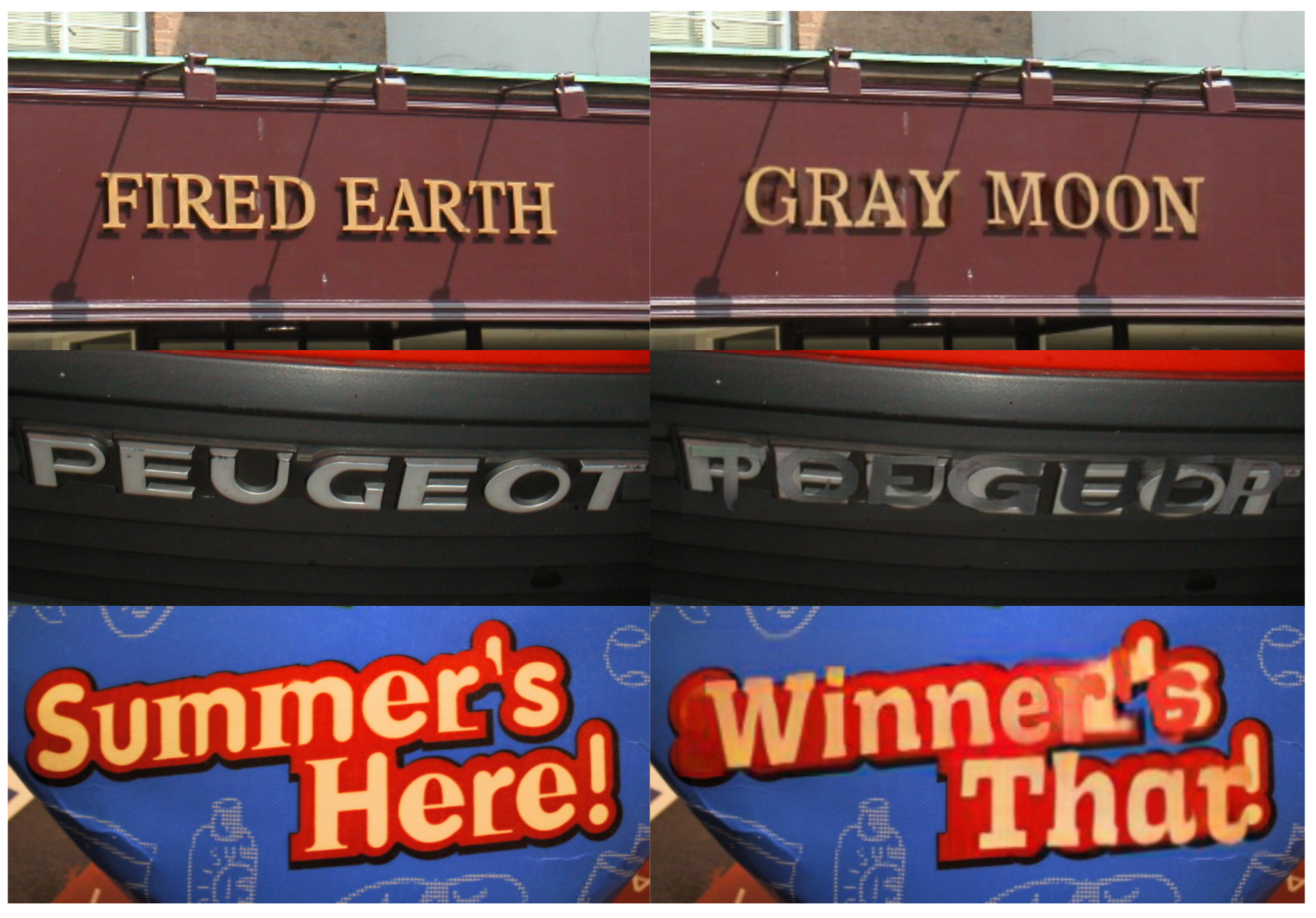}

\caption{The failure cases. Left: source images; right: edited results.}
\label{fig:fail}
\vspace{-0.5cm}
\end{figure}

\subsection{Failure Cases}
Although our method is capable of most scene images, there are still some limitations. Our methods may fail when the text have very complex structures or rare font shapes. Fig.~\ref{fig:fail} shows some failed cases of our method. In the top row, although the foreground text has been transferred successfully, it can be found that the shadow of the original text still remains in the output image. In the middle row of images, our model fails to extract the style of text with such a complicated spatial structure, and the result of the background erasure is also sub-optimal. In the bottom row of images, the boundaries surrounding the text are not transfered with text. 
We attribute these failure cases to the inadequacy of these samples in training data, so we assume they could be alleviated by augmenting the training set with more font effects.

\section{Conclusion and Future Work}
This paper proposes an end-to-end network for text editing task, which can replace the text from scene text image while maintaining the original style. We mainly divide it into three steps to achieve this function: (1) extract foreground text style and transfer to input text with the help of skeleton; (2) erase the style image with appropriate texture to get background image; (3) merge the transferred text with the erased background.
To our best knowledge, this paper is the first work to edit text image in the word-level. Our method has achieved outstanding results in both subjective visual realness and objective quantitative scores on ICDAR13 dataset. At the same time, the network also have the ability to erase text and edit on cross-language situation, and the effectiveness of our network has been verified through the comprehensive ablation studies.

In the future, we hope to solve text editing in more complex scenarios while making the model easier to use. We will edit text between more language pairs to fully exploit the ability of the proposed model. We will try to propose new evaluation metrics to evaluate the quality of text editing properly. 

\begin{acks}
This work is supported by NSFC 61733007, to Dr. Xiang Bai by the National Program for Support of Top-notch Young Professionals and the Program for HUST Academic Frontier Youth Team 2017QYTD08. We sincerely thank Zhen Zhu and Tengteng Huang for their valuable discussions and continuous help to this paper.
\end{acks}

%
\bibliographystyle{ACM-Reference-Format}
\bibliography{SRNet}

%

\end{document}